\documentclass{article}

\usepackage{PRIMEarxiv}

\usepackage[utf8]{inputenc} 
\usepackage[T1]{fontenc}    
\usepackage{hyperref}       
\usepackage{url}            
\usepackage{booktabs}       
\usepackage{amsfonts}       
\usepackage{nicefrac}       
\usepackage{microtype}      
\usepackage{lipsum}
\usepackage{fancyhdr}       
\usepackage{graphicx}       
\graphicspath{{media/}}     

\title{Creating a Silver Standard for Patent Simplification
\thanks{\textit{\underline{Citation}}: 
\textbf{Silvia Casola, Alberto Lavelli, and Horacio Saggion. 2023. Creating a Silver
Standard for Patent Simplification. In Proceedings of the 46th International
ACM SIGIR Conference on Research and Development in Information Retrieval
(SIGIR ’23), July 23–27, 2023, Taipei, Taiwan. ACM, New York, NY, USA,
11 pages. https://doi.org/10.1145/3539618.3591657
}} 
}

\author{
  Silvia Casola \\
  University of Padua \\
  Fondazione Bruno Kessler \\
  Italy \\
  \texttt{scasola@fbk.eu} \\
   \And
  Alberto Lavelli \\
  Fondazione Bruno Kessler \\
  Italy \\
  \texttt{lavelli@fbk.eu} \\
   \And
   Horacio Saggion \\
  Universitat Pompeu Fabra \\
  Spain \\
\texttt{horacio.saggion@upf.edu} \\
}

\usepackage{cancel}
\usepackage{placeins}
\usepackage{amsmath}
\begin{document}
\maketitle

\begin{abstract}
Patents are legal documents that aim at protecting inventions on the one hand and at making technical knowledge circulate on the other.
    Their complex style -- a mix of legal, technical, and extremely vague language -- makes their content hard to access for humans and machines and poses substantial challenges to the information retrieval community. This paper proposes an approach to automatically simplify patent text through rephrasing. Since no in-domain parallel simplification data exist, we propose a method to automatically generate a large-scale silver standard for patent sentences. To obtain candidates, we use a general-domain paraphrasing system; however, the process is error-prone and difficult to control. Thus, we pair it with proper filters and construct a cleaner corpus that can successfully be used to train a simplification system. Human evaluation of the synthetic silver corpus shows that it is considered grammatical, adequate, and contains simple sentences.
    \end{abstract}

\keywords{text simplification \and natural language processing \and patents}

\section{Introduction}
To file a patent, an inventor must describe their invention in detail so that a person skilled in the art\footnote{A person skilled in the art has ordinary skills in a particular technical field.} can understand and reproduce it. The invention disclosure should benefit society by making innovation circulate. 
However, patent documents are extremely complex: they contain long sentences, novel multiterm entities, and complex syntax built out of noun phrases instead of clauses with recurring entities \cite{my_survey}. \cite{evalSim} performed a user study on patent text readability: 
Most participants strongly (35\%) or somewhat (23\%) agreed improving the readability of patents was important and considered documents extremely (29\%) or somewhat (29\%) difﬁcult to read. 39\% of the participants reported difﬁculties in ﬁnding information they were looking for.
Even law scholars have advocated for simpler language in patent documents. \cite{plain} argues that "When the subject of the case is wrapped in complex and unfamiliar terms, it is tremendously difficult for legal actors to grapple with the theoretical content of the dispute. [...] 
Communication at the intersection of law and science will always be tremendously challenging. Nevertheless, there are elements of the current patent system that substantially exacerbate the problem". Moreover, patent text is difficult to process automatically and is significantly challenging for the information retrieval community.

Previous work on patent simplification has mainly targeted experts: systems search ways to help patent attorneys better understand the claims (which defines the scope of the legal protection) through ad hoc visualizations or compare documents in the same patent family\footnote{A patent family is a collection of patent applications covering the same or similar technical content.}. Since the legal scope needs to remain unchanged, modifying the text presentation is preferred to rephrasing.
We take a different stance and aim at creating a silver standard for simplifying the Description of the invention -- the section that describes the invention embodiments in detail -- through rephrasing. Simplifying the Description makes the technical knowledge more accessible to society, practitioners in the field, engineers, academics, and other laypeople; moreover, it can improve performance on further automatic downstream tasks, as shown in other domains \cite{simpl_ie}.

The state of the art in sentence simplification through rephrasing relies on parallel corpora of complex and simple sentences. The problem is framed as a monolingual translation: using a sequence-to-sequence network, complex sentences get translated into their corresponding simple versions. 
However, parallel data for simplification are difficult to obtain, and many of the existing large-scale datasets derive from the automatic alignment of sentences in the English and Simple English Wikipedia \cite{wikismall, wikilarge}. Creating parallel datasets in other domains requires substantial human effort and is slow and expensive. 
To the best of our knowledge, no simplified corpora exist for the patent domain. 

This paper proposes a method to automatically create a parallel simplification corpus of patent sentences. In particular, we generate a huge noisy corpus of simplification candidate pairs (that we call "bronze'') and clean it to obtain a higher-quality silver corpus. To generate the candidates, we adopt a paraphrasing system trained on general-domain text and show that using a zero-shot approach on patents results in simpler, shorter, and easier-to-read sentences. The process is, however, hard to control and might be error-prone. Thus, we discuss filters to select reliable candidates only and show how the silver corpus can be used for training controllable simplification systems for patent sentences. 
Our contributions are the following:
\begin{enumerate}
    \item We propose the use of a paraphrasing system to obtain simplification candidates from complex patent sentences. The paraphraser is used in a zero-shot fashion and is only trained on general-domain sentences. We show that, while far from perfect, this zero-shot approach produces text which is more compressed and thus simpler to read. We call the simplifications obtained this way a "bronze corpus".
    \item We discuss filtering to select candidates that are appropriate for simplification only. By doing so, we generate the first large-scale parallel silver standard for patent sentence simplification. We make the bronze and silver parallel datasets public for future research\footnote{The corpora and models are available at \url{https://github.com/slvcsl/patentSilverStandard}. Data is shared under a Creative Commons Attribution-NonCommercial-ShareAlike 3.0 Unported License.}.
    \item We show that the silver standard can be used to train a sequence-to-sequence state-of-the-art system for controllable patent simplification that we release.
    \item Finally, we perform a human evaluation of the results and make the unaggregated data public. During the evaluation, we also collect non-expert-generated simplification, which can be used in future research.
\end{enumerate}

\section{Previous work}
\label{sec:prev_work}
\subsection{Sentence simplification}
Automatic text simplification aims at making text easier to read and understand for a target group (e.g., people who are neurodivergent \cite{simpl_autism} or have a  disability \cite{disability} or people with a low literacy level \cite{non_native}). Most of the current research performs simplification at the sentence level; refer to  \cite{alva-manchego-etal-2020-data} for a survey. To simplify complex sentences, the applied transformations  can be lexical \cite{lexical} (substituting unusual words or explaining hard concepts) or syntactic (with sentence splitting \cite{splitting} and other syntactic structure simplification techniques). They can involve removing unimportant information and compressing the text to enhance clarity.   
The most popular paradigm for sentence simplification through rephrasing employs sequence-to-sequence architectures. The task is framed as a translation between complex and simple sentences \cite{seq2seq_simpl}. Controlling the level of simplification (e.g., in terms of compression or extent of the rephrasing) is now deemed essential, and ad hoc control tokens are often used for this goal. \cite{access}, for example, proposed to enhance a transformer \cite{transformer} with control tokens to guide the level of compression, similarity to the input, and lexical and syntactic complexity.  \cite{kim_t5-controllable} used a T5 model \cite{t5} and proposed a new control token to help the model replace long, complex words with shorter ones.

These architectures require a large number of parallel complex and simple sentences to train. A popular dataset for simplification is Wikilarge \cite{wikilarge}, obtained by automatically aligning sentences from corresponding documents in the English and Simple English Wikipedia. 
However, obtaining corpora in other domains and languages is more complex, as no natural alignment between documents usually exists. Only a few manually curated datasets have been created (e.g., the Newsela corpus \cite{newsela}), as they imply a huge human effort, which is expensive and requires considerable time. 
Researchers have thus proposed ways to automatically generate silver standards for simplification. Starting from a Japanese corpus of mixed complexity, \cite{jap_pseudo_parallel} first identified a complex and a simple subcorpus based on readability scores; then, they aligned sentences from the two corpora based on their word embeddings nearest neighbors; they  experimented with several  similarity metrics. 
Similarly, \cite{martin-paraphrase} used the LASER \cite{laser} sentence embedding and retrieved nearest neighbors with some filters to ensure quality. While interesting, these methods are hard to apply to the patent domain, whose textual content has largely the same (high) complexity, as strict rules and common patterns govern the style. 
In contrast, \cite{simplif-backtranslation} proposed a method to turn a translation corpus into a simplification corpus. Given a pair of sentences in two languages, one is used as a bridge and translated into the other (target) language. Authors argue that the two sentences will likely have a different complexity level because machine translation models tend to output high-frequency tokens \cite{high_freq_mt}, and there is often a difference in complexity between languages in translation corpora \cite{complexity_mt_corpora}. If the translation is satisfactory, and if there is a difference in the complexity level, the sentence pair is added to the simplification silver standard. Authors show that using a large filtered silver corpus obtained this way outperforms a smaller, cleaner corpus. This approach shifts the burden from a parallel simplification corpus to a parallel translation corpus, which is typically easier to obtain. All described methods were proposed for general-domain simplification. 

Our method is inspired by previous literature on generating silver data for simplification in that we propose to select relevant pairs from a larger corpus of possible candidates. However, we do not rely on in-domain simple data nor on external parallel data.

\subsection{Patent simplification}
Patent documents have attracted great interest from the Information Retrieval community, as demonstrated, e.g., by the CLEF-IP labs \cite{clef2009, clef2010, clef2011, clef2012, clef2013}. Refer to \cite{shalaby2019patent} and \cite{lupu2013patent} for recent surveys in patent retrieval and to \cite{my_survey} for a review of previous work on patent simplification.
As claims are hard to read even for patent professionals, most previous work has focused on improving the accessibility and readability of the Claim section. 
\cite{ferraro}, for example, aimed at improving each claim presentation by segmenting it into fragments that are then formatted more clearly, e.g., by adding new lines. \cite{informationExtraction} used an Information Extraction engine that detects entity mentions, their type, and relations through distant supervision. They built an interface that shows the most salient elements in the Claim section to understand the patent structure and compare patents in the same family. These works target patent attorneys and other experts involved in the patent filing process. In contrast, \cite{evalSim} proposed ways to improve patent visualization to lay people, all of which were considered at least as good as the original patent text by users. Some previous research has also tried to improve the understanding of entities in the claims by linking them to the Description, where they are mentioned in the context of actual embodiments \cite{jap}.
Finally, other previous work has visualized claims in a more structured way, e.g., through graphs \cite{lupu} or trees \cite{multiple}.
Previous work on simplification through rephrasing is much more limited. A rewriting and rephrasing system was built as part of the PATExpert project~\cite{patexpert}. Researchers considered
two levels of simplification: one uses surface criteria to segment the input and reconstructs chunks into shorter, easier-to-read sentences ~\cite{simplificationShallow}. The other ~\cite{accessible} represented patents by their Deep Syntactic Structures. This representation is, in turn, used to rewrite a text that is simpler to process for the
reader (possibly in another language). Both methods modify the patent text.

\section{Method}
\label{sec:method}


\subsection{Dataset}
\label{sec:dataset}
We use data in the Patent Translation Resource (PatTR) \cite{pattr, pattr2} corpus. PatTR is a sentence-level parallel corpus extracted from the MAREC patent collection\footnote{See: \url{http://www.ifs.tuwien.ac.at/imp/marec.shtml} [Last accessed January 2023]}. The data is available under a  Creative Commons Attribution-NonCommercial-ShareAlike 3.0 Unported License. It consists of sentence pairs in different languages (DE-EN, FR-EN, and DE-FR). 
Sentence pairs are indexed by language and patent sections, i.e., Title, Abstract, Description, and Claims. 
We chose the PaTR as it contains pre-tokenized sentences, and finding sentence delimiters is an error-prone task for patents. Moreover, the PaTR dataset allows us to compare our method with the general-domain approach by \cite{simplif-backtranslation}, which requires a parallel translation corpus.

In the following experiments, we will focus on the English sentences from the DE-EN pairs and specifically on sentences extracted from the Description only. 
 The PatTR DE-EN Description dataset contains almost 12 million sentence pairs. For computational reasons, we sample 500 thousand sentences. 
 
 As a preprocessing step, we removed sentences shorter than 5 tokens or longer than 55 tokens. We also filter sentences where alphabetic characters account for less than 60\% of the total. These sentences mainly contain long lists of references or complex chemical formulas. This leaves us with 425,148 sentences.
Finally,  sentences in the Description often contain references to the figures in the form of numbers in brackets. We use regular expressions to remove these references. 

Table~\ref{tab:stats_sentences} contains some statistics on the English subset of the corpus. 
We use the following metrics as simplicity proxies:
\begin{itemize}
    \item Flesch Reading Ease score \cite{fre}: it computes sentence simplicity as a function of the mean word length (in syllables) and the mean sentence length (in words). A higher score indicates a text that is easier to read. 
    \item Flesch Kincaid Grade Level \cite{fkgl}: it is similar in nature to the Flesch Reading Ease score, but it is weighted to correspond to U.S. education grade levels. We report it for completeness. A lower score indicates a text that is easier to read.
    \item WordRank score: this measure uses  word frequency as a proxy of lexical simplicity (as text containing more common words is considered easier to understand). It is computed by taking the third quartile of log ranks (inverse frequency order) of all words in the sentence. We use the implementation proposed by \cite{access}. 
    \item Maximum Dependency Tree: this score uses the height of the dependency tree as a proxy for syntactic complexity. We use the implementation proposed by \cite{access}.
\end{itemize}
All metrics are computed per sentence and then averaged.

\begin{table}[]
  \caption{Statistics on the original English
  patent sentences (after preliminary filtering)}
  \label{tab:stats_sentences}
    \centering
    \begin{tabular}{lc}
    \toprule
    Metric & Mean $\pm$ std\\
    \midrule
    Flesch Reading Ease \cite{fre} & 32.5 $\pm$ 26.3 \\
    Flesch–Kincaid Grade Level \cite{fkgl} & 61.1 $\pm$ 9.5 \\
    WordRank \cite{access} & 9.8 $\pm$ 1.1 \\
    Max dependency tree depth & 6.7 $\pm$ 2.5 \\
    Length (chars) & 170.3 $\pm$ 75.3 \\   
    \bottomrule
    \end{tabular}%
    \end{table}

\subsection{Generating simplification candidates using a general-domain paraphrasing system}
Previous work has explored using models trained on out-of-domain simple text to simplify complex text. \cite{surya-unsupervised}, for example, propose an architecture with two decoders (one trained on complex text only and the other trained on simple text only) to control the level of simplification without supervision.
Inspired by this line of work, we investigate using a sequence-to-sequence system trained on general-domain text only for simplifying complex patent text.
Specifically, we employ a Pegasus-based \cite{pegasus} paraphrasing system for general text. The system was fine-tuned on a custom set of 60k examples from  multiple datasets, including PAWS \cite{paws}. The trained model is available on Huggingface\footnote{\url{tuner007/pegasus_paraphrase} [Last accessed: January 2023]}. We use it as a black box and transform the preprocessed patent sentences.  

Table~\ref{tab:results_pegasus} reports some random sentence pairs (no cherry-picking) generated using Pegasus. 
The simplification is mainly by sentence compression and lexical substitution; some other syntactical changes (e.g., use of the active voice instead of the passive voice) are also common. 

While the generated candidates tend to be simpler than the original sentences, using the Pegasus model directly for simplification has two main limitations:
\begin{enumerate}
    \item Some of the generated sentences contain errors and unknown tokens or are too similar or excessively compressed with respect to the original sentence.
    \item The process is hard to control. While state-of-the-art simplification systems allow  controlling the level of compression through appropriate tokens, the Pegasus model has a strong tendency to compression. However, this tendency can degenerate into text that does not contain the original core elements, lose too much content, or completely change the original meaning. The level of paraphrasing, similarity, and lexical simplification is also hard to control.
\end{enumerate}

Due to these issues, in the following, we will call the candidates generated by our Pegasus model our "bronze corpus".

\begin{table*}[]
\caption{Simplification candidate pairs obtained through the Pegasus paraphrasing model (zero-shot) prior to any filtering. Pairs with a crossed-out number are later filtered.}
\label{tab:results_pegasus}
    \centering
    \begin{tabular}{lll}
    \toprule
     & \footnotesize{Original} & \footnotesize{Candidate} \\
    \midrule
     \cancel{0} & \parbox{8cm}{\footnotesize{In interaction with the component secured on the cylinder head, the radial shoulder of the pressure medium distributor now prevents the camshaft from migrating axially further into the cylinder head.}} & \parbox{8cm}{\footnotesize{The radial shoulder of the pressure medium distributor prevents the camshaft from moving further into the cylinder head when interacting with the component secured on the cylinder head.}}\\
     1 & \parbox{8cm}{\footnotesize{Film 12 had holes, and these could be attributed to moisture in the blend material.}} & \parbox{8cm}{\footnotesize{The blend material could be to blame for the holes in Film 12.}}\\
     \cancel{2} & \parbox{8cm}{\footnotesize{According to the invention, the formulations advantageously comprise one or more antioxidants.}} & \parbox{8cm}{\footnotesize{The invention states that the formulations comprise one or more anti-oxidants.}}\\
     3 & \parbox{8cm}{\footnotesize{In the pressure medium flow path between the consumers and the variable displacement pump a respective meter-in orifice is provided.}} & \parbox{8cm}{\footnotesize{A meter-in orifice is provided in the pressure medium flow path between the consumers and the pump.}}\\
     \cancel{4} & \parbox{8cm}{\footnotesize{The yield of TMCH relative to the feed KIP was 27\%.}} & \parbox{8cm}{\footnotesize{The yield was 27\%.}}\\
     5 & \parbox{8cm}{\footnotesize{A throttle valve angle sensor for internal combustion engines which is described therein has a permanent magnet that is attached to one end of the throttle valve shaft and produces a constant magnetic flux having a direction which is turned as a function of the rotation of the throttle valve shaft.}} & \parbox{8cm}{\footnotesize{A permanent magnet that is attached to one end of the throttle valve shaft produces a constant magnetic flux which is turned as a function of the rotation of the throttle valve shaft.}}\\
     6 & \parbox{8cm}{\footnotesize{They have to be so thin in comparison with the wavelengths produced that they may be caused to vibrate by the ultrasonic waves.}} & \parbox{8cm}{\footnotesize{They have to be so thin that they can't be seen by the naked eye}}\\
     \cancel{7} & \parbox{8cm}{\footnotesize{Mechanisms are provided which allow the guide rollers to yield under spring tension so that the force with which the rollers are pressed on the pane is held within a prescribed range regardless of installation and vehicle body tolerances.}} & \parbox{8cm}{\footnotesize{The force with which the rollers are pressed on the pane is held within a prescribed range, regardless of installation and vehicle body tolerances, because of mechanisms provided which allow the guide rollers to yield under spring tension.}}\\
     8 & \parbox{8cm}{\footnotesize{The line includes an outer conductor tube which is equipped on both ends with connection flanges .}} & \parbox{8cm}{\footnotesize{The line has an outer conductor tube on both ends.}}\\
     9 & \parbox{8cm}{\footnotesize{Thick pressure-sensitive adhesive coats may be present in unfilled, all-acrylate form or resin-blended form or filled with organic or inorganic fillers.}} & \parbox{8cm}{\footnotesize{Thick pressure-sensitive adhesive coats can be filled with organic or inorganic fillers.}}\\
     \cancel{10} & \parbox{8cm}{\footnotesize{The seamless tubular casing preferably has a thickness of from 40 to 200 \textbackslash u03bcm, when it is made by a blown tube process, and a thickness of from 25 to 75 \textbackslash u03bcm, when it was obtained by biaxial orientation (double bubble process).}} & \parbox{8cm}{\footnotesize{When it is made using a blown tube process and a thickness of 25 to 75?m is obtained, the seamless tubular casing should have a thickness of 40 to 200?m}}\\
     11 & \parbox{8cm}{\footnotesize{Furthermore, with this type of integration it is mainly the classic MC functionalities, as are relevant in particular for machine tools, that are supported.}} & \parbox{8cm}{\footnotesize{The classic MC functionalities are relevant in particular for machine tools that are supported by this type of integration.}}\\
     \cancel{12} & \parbox{8cm}{\footnotesize{Then, if a predetermined limiting pressure value is reached, the valve body member 19 is moved against the pretension force of the spring member 18, thus clearing a further conduit with respect to the liquid treatment apparatus 3.}} & \parbox{8cm}{\footnotesize{If a limiting pressure value is reached, the valve body member 19 is moved against the pretension force of the spring member 18, thus clearing a further conduit with respect to the liquid treatment apparatus 3.}}\\
     13 & \parbox{8cm}{\footnotesize{These process parameters for intermingling depend on the yarn titer and the lubricant applied to the yarn and must therefore be adapted accordingly.}} & \parbox{8cm}{\footnotesize{The process parameters for intermingling are dependent on the yarn titer and the lubricant applied to the yarn.}}\\
     14 & \parbox{8cm}{\footnotesize{The telephone network is formed as a digital telephone network (e.g. Integrated Services Digital Network, ISDN) or as an analogue telephone network (e.g. Public Switched Telephone Network, PSTN).}} & \parbox{8cm}{\footnotesize{The telephone network can be formed as a digital telephone network or an analogue telephone network.}}\\
     \cancel{15} & \parbox{8cm}{\footnotesize{This may be perceived as disadvantageous by the consumer.}} & \parbox{8cm}{\footnotesize{This may be seen as disadvantageous by the consumer.}}\\
     16 & \parbox{8cm}{\footnotesize{On the other end of the coupling rods 20, 21, these are connected by way of second ball-and-socket joints 22, 23 as the second connection joint of the parallelogram linkage with the second axis 24 which extends in parallel to the first axis 19.}} & \parbox{8cm}{\footnotesize{The second connection joint of the parallelogram linkage with the second axis 24 extends in parallel to the first axis 19 on the other end of the coupling rods.}}\\
     \bottomrule
    \end{tabular}%
\end{table*}

For comparison, we also report the same candidates obtained through translation (prior to any filtering) using the model proposed by \cite{simplif-backtranslation} in Table \ref{tab:simpl_transl_all}. Note that the sentences are practically paraphrases, and there is no evident difference in complexity. We attribute these characteristics to the lack of complexity differences in sentences in different languages in the patent domain. Our preliminary experiments show that even after filtering these instances for simplification level, e.g., maintaining only pairs with a high relative difference in reading scores, as the original method proposes, the level of simplification is poor.

\begin{table*}[]
\caption{Pair candidates obtained through translation, as proposed by \cite{simplif-backtranslation}.}
\label{tab:simpl_transl_all}
\centering
\begin{tabular}{lll}

    \toprule
     & \footnotesize{Original} & \footnotesize{Candidate} \\
    \midrule
     0 & \parbox{8cm}{\footnotesize{In interaction with the component secured on the cylinder head, the radial shoulder of the pressure medium distributor now prevents the camshaft from migrating axially further into the cylinder head.}} & \parbox{8cm}{\footnotesize{In this way, the radial collar of the distributor forms the axial bearing of the camshaft in conjunction with the cylinder-head-resistant component.}}\\
     1 & \parbox{8cm}{\footnotesize{Film 12 had holes, and these could be attributed to moisture in the blend material.}} & \parbox{8cm}{\footnotesize{Foil 12 had holes that could be traced back to moisture in the glare material.}}\\
     2 & \parbox{8cm}{\footnotesize{According to the invention, the formulations advantageously comprise one or more antioxidants.}} & \parbox{8cm}{\footnotesize{According to the invention, the preparations contain advantageous one or more antioxidants.}}\\
     3 & \parbox{8cm}{\footnotesize{In the pressure medium flow path between the consumers and the variable displacement pump a respective meter-in orifice is provided.}} & \parbox{8cm}{\footnotesize{A diaphragm is provided in each case in the pressure medium flow path between the consumers and the variable displacement pump.}}\\
     4 & \parbox{8cm}{\footnotesize{The yield of TMCH relative to the feed KIP was 27\%.}} & \parbox{8cm}{\footnotesize{The yield of THCM in relation to the KIP used was 27\%.}}\\
     5 & \parbox{8cm}{\footnotesize{A throttle valve angle sensor for internal combustion engines which is described therein has a permanent magnet that is attached to one end of the throttle valve shaft and produces a constant magnetic flux having a direction which is turned as a function of the rotation of the throttle valve shaft.}} & \parbox{8cm}{\footnotesize{Here a throttle angle sensor for internal combustion engines is described with a permanent magnet attached to one end of the throttle shaft, which generates a constant magnetic flux, the direction of which is rotated depending on the rotation of the throttle shaft.}}\\
     6 & \parbox{8cm}{\footnotesize{They have to be so thin in comparison with the wavelengths produced that they may be caused to vibrate by the ultrasonic waves.}} & \parbox{8cm}{\footnotesize{These must be so thin in comparison to the wavelengths produced that they can be oscillated by the ultrasonic waves.}}\\
     7 & \parbox{8cm}{\footnotesize{Mechanisms are provided which allow the guide rollers to yield under spring tension so that the force with which the rollers are pressed on the pane is held within a prescribed range regardless of installation and vehicle body tolerances.}} & \parbox{8cm}{\footnotesize{In order to ensure that the force with which rollers 56 are pressed to disc 4 remains within a predetermined range, independent of installation and body tolerances, means 42.70 are provided to enable guide rollers 56 to be cushioned.}}\\
     8 & \parbox{8cm}{\footnotesize{The line includes an outer conductor tube which is equipped on both ends with connection flanges .}} & \parbox{8cm}{\footnotesize{The cable comprises an outer conductor tube 1, which is equipped with connecting flanges 2 at both ends.}}\\
     9 & \parbox{8cm}{\footnotesize{Thick pressure-sensitive adhesive coats may be present in unfilled, all-acrylate form or resin-blended form or filled with organic or inorganic fillers.}} & \parbox{8cm}{\footnotesize{Thick adhesive coatings can be available unfilled as pure acrylate or mixed with resin or filled with organic or inorganic fillers.}}\\
     10 & \parbox{8cm}{\footnotesize{The seamless tubular casing preferably has a thickness of from 40 to 200 \textbackslash u03bcm, when it is made by a blown tube process, and a thickness of from 25 to 75 \textbackslash u03bcm, when it was obtained by biaxial orientation (double bubble process).}} & \parbox{8cm}{\footnotesize{The seamless, tubular sheath of the invention preferably has a thickness of 40 to 200 \textbackslash u00b5m if manufactured by a tube blowing process, and a thickness of 25 to 75 \textbackslash u00b5m if obtained by biaxial stretch orientation.}}\\
     11 & \parbox{8cm}{\footnotesize{Furthermore, with this type of integration it is mainly the classic MC functionalities, as are relevant in particular for machine tools, that are supported.}} & \parbox{8cm}{\footnotesize{In addition, this type of integration mainly supports the classic MC functionalities, which are particularly relevant for machine tools.}}\\
     12 & \parbox{8cm}{\footnotesize{Then, if a predetermined limiting pressure value is reached, the valve body member 19 is moved against the pretension force of the spring member 18, thus clearing a further conduit with respect to the liquid treatment apparatus 3.}} & \parbox{8cm}{\footnotesize{If a certain limit value is exceeded, valve body 19 is shifted against the prestressing force of spring 18, thus releasing a further line with respect to liquid treatment device 3.}}\\
     13 & \parbox{8cm}{\footnotesize{These process parameters for intermingling depend on the yarn titer and the lubricant applied to the yarn and must therefore be adapted accordingly.}} & \parbox{8cm}{\footnotesize{These process conditions for fishing depend on the yarn titer and the preparation applied to the yarn and must therefore be adapted to these parameters.}}\\
     14 & \parbox{8cm}{\footnotesize{The telephone network is formed as a digital telephone network (e.g. Integrated Services Digital Network, ISDN) or as an analogue telephone network (e.g. Public Switched Telephone Network, PSTN).}} & \parbox{8cm}{\footnotesize{Telephone Network 2 is designed as a digital telephone network (e.g. Integrated Services Digital Network, ISDN) or an analog telephone network (e.g. Public Switched Telephone Network, PSTN).}}\\
     15 & \parbox{8cm}{\footnotesize{This may be perceived as disadvantageous by the consumer.}} & \parbox{8cm}{\footnotesize{This can be detrimental to the consumer.}}\\
     16 & \parbox{8cm}{\footnotesize{On the other end of the coupling rods 20, 21, these are connected by way of second ball-and-socket joints 22, 23 as the second connection joint of the parallelogram linkage with the second axis 24 which extends in parallel to the first axis 19.}} & \parbox{8cm}{\footnotesize{At the other end of the coupling rods 20, 21, these are connected via second ball joints 22, 23, as the second joint of the parallelogram guide, to the second axis 24, which runs parallel to the first axis 19.}}\\
     \bottomrule
    \end{tabular}%
\end{table*}

\subsection{Filtering bad candidates and generating a silver standard of patent sentences.}

As the bronze corpus is noisy, we remove candidates according to the following criteria:
\begin{itemize}
    \item Bad tokens: we remove any pair that contains an unknown token. We also noticed that the generation can degenerate into a sequence of numerical tokens (e.g., 65561). We remove any pair where the generated sentence contained any of these tokens. For example, this removes pair 10 in Table~\ref{tab:results_pegasus}.
    \item Non-alphabetical characters: we remove pairs where the alphabetical characters in the generated sentence account for less than 60\% of the total. These sentences are not necessarily the effect of a bad generation, but the pair is useless as a training example.  
    \item Similarity: we remove sentences whose character-level Levenshteins similarity between the original sentences and its simplification (normalized by the sum of their length) is < 25\% or > 90\%. The low threshold removes simple sentences that are too different compared to the complex ones (generally very short); the high threshold removes sentences that are equal to the original one or only differ by a small number of characters. This removes, for example, pair 12 and 15. We also remove pairs in which the generated sentence has a similarity > 99\% with a continuous substring of the original sentence. Finally, we removed pairs where the strings obtained by sorting the tokens have a similarity > 90\%. This removes, for example,  pairs 0, 2, and 7.
    \item Compression: we remove pairs where the ratio between the generated and the original sentence length was > 1.5 or < 0.5. This step avoids excessive compression, which generally corresponds to losing important content or having a modified meaning. This removes, for example, pair 4.
    \item Simplicity: 
    we considered the Fresh Reading Ease score, the WordRank score, and the sentence dependency tree height as simplification proxies. We removed candidates where the generated sentence is not simpler according to any of these scores.
\end{itemize}

\begin{table*}[]

  \caption{Number of instances removed by each filter from our original 426,963 bronze standard and examples of such instances. The filters were applied consecutively.}
  \label{tab:filters}
    \centering
    \begin{tabular}{llll}
    \toprule
    Filter & Removed & Examples \\
    \midrule
    Bad tokens & 8,602 & \parbox{12cm}{\footnotesize{\textbf{Simple:} \textit{"It is1-65561-65561-65561-65561-65561-65561-65561-65561-65561-65561-65561-65561-65561-65561-65561-65561-65561-65561-65561-[...]-65561-65561-65561-6556"}}} \\
    \hline
    Non-alphabetical & 695 &  \parbox{10cm}{\footnotesize{\textbf{Simple:} \textit{"R.sup.1, R.sup.2, R.sup.3, Z.sup.1, Z.sup.2 and Z.sup.3 are all related."}}} \\
    \hline
    Similarity & 38,217 & \parbox{12cm}{\footnotesize{\textbf{Complex:} \textit{"Example 23 (E,E)-3,3-[4-(4-Fluorophenyl)-6-isopropyl-1-methyl-pyrid-2one-3,5-diyl]-diprop-2-enal \#\#STR54\#\# Analogously to Example 10, 2.70 g of crude product are obtained starting from 2.13 g (7.1 mmol) of the compound of Example 22."} \textbf{Simple:} \textit{"There are 2.70 g of crude product."} \\
    \textbf{Complex:} \textit{"Another method of solution attempts to achieve optimum operating cracks by means of temperature-dependent shifting of a conical rotor in the tapered case diameter with the help of an adjustment device."} \textbf{Simple:} \textit{"Another method of solution attempts to achieve optimum operating cracks by means of temperature dependent shifting of a conical rotor in the tapered case with the help of an adjustment device."}}} \\
    \hline
    Partial similarity & 5,937 & \parbox{12cm}{\footnotesize{\textbf{Complex:} \textit{"In the treatment of parts of plants, the active compound concentrations in the use forms can be varied within a substantial range."} \textbf{Simple:} \textit{"The active compound concentrations in the use forms can be varied within a substantial range."}}} \\
    \hline
    Sorted similarity & 18,807 & \parbox{12cm}{\footnotesize{\textbf{Complex:} \textit{"The solution polymerization without addition of other auxiliaries is the preferred process for the ethylene-vinyl acetate copolymers to be used according to the present invention."} \textbf{Simple:} \textit{"According to the present invention, the solution polymerization without addition of other auxiliaries is the preferred method for the production of the ethylene-vinyl acetate copolymers."}}} \\
    \hline
    Compression & 62,926 & \parbox{12cm}{\footnotesize{\textbf{Complex:} \textit{"he bottom end of this rod passes through a hole 46 which leads into the fourth circular chamber 20."} \textbf{Simple:} \textit{"The hole 46 leads into the fourth chamber 20."}
    }} \\
    \hline
    Simplicity & 1,999 & \parbox{12cm}{\footnotesize{\textbf{Complex:} \textit{"The preparation and further processing of the catalyst supports used according to the invention are well known to the person skilled in the art."} \textbf{Simple:} \textit{"The person skilled in the art knows about the preparation and further processing of the catalyst supports used in the invention."}
    }} \\
    \bottomrule
    \end{tabular}%
    \end{table*}

Table~\ref{tab:filters} shows the number of pairs removed by each filtering step, with examples.  
We chose the thresholds heuristically. Future work could investigate learning their most suitable values from a larger corpus of annotated silver sentences; moreover, it could study the best threshold between noise and size of the corpus.
The remaining sentences compose our parallel silver standard. After filtering, our corpus consists of 287,965 samples.

\subsection{Using the corpus for training a controllable simplification system}
A silver standard for simplification allows for experimentation with models and training processes.
To demonstrate this point, we train ACCESS \cite{access}, a state-of-the-art system for automatic sentence simplification. 
We randomly split the silver standard into train (184,297), validation (46,075), and test (57,593) sets. 
We train the model with early stopping on the validation SARI \cite{sari} (patience: 20).
We also evaluate the model on the human annotations we collected from the Prolific workers (Section \ref{sec:human_annotation}).
 
Table~\ref{tab:results_training} shows the results, as computed using the EASSE library \cite{easse}. 
As expected, the metrics on the silver standard are particularly high. This is likely because the corpus is synthetic and presents patterns that are easy to learn for the simplification model. 
Note, however, that when the model is evaluated on human-written sentences, its performance is close to those obtained on other general-domain datasets. Specifically, we have reported the results on  the human evaluation set when using it both as a validation set (allowing the ACCESS model to optimize its preprocessing parameters) and as a test set (only using the optimal parameters computed on the silver standard validation set). In future work, we plan to compare our results with those obtained by training sequence-to-sequence models on different data sets.

\begin{table}[]
  \caption{Results of the model trained using the silver standard}
\label{tab:results_training}
    \centering
    \begin{tabular}{lll}
    \toprule
     & SARI & BLEU\\
    \midrule
    Validation (silver) & 55.09 & 54.88 \\
     Test (silver) & 55.22 & 54.99 \\
     Human simplification (with param. search) & 39.25 & 56.86  \\
     Human simplification (without param. search) & 36.99 & 57.78  \\
     \bottomrule
     \end{tabular}%
\end{table}

Having a silver parallel corpus also allows experimenting with transfer learning. We plan to explore the impact of consecutive fine-tunings (for example, further fine-tuning a model already fine-tuned on Wikipedia-style data). On the other hand, the performance of a model first fine-tuned on a large-scale silver dataset can usually be improved with a small human-curated dataset; this approach would diminish the human effort and, thus, the time and money required for obtaining high-quality simplifications. We plan to investigate these aspects in future work. 

\section{Corpus quality estimation}
\label{sec:eval_corpus}
\subsection{Automatic metrics}
\begin{figure*}[]
\centering
\begin{minipage}[b]{0.32\textwidth}
\centering
\includegraphics[width=\textwidth]{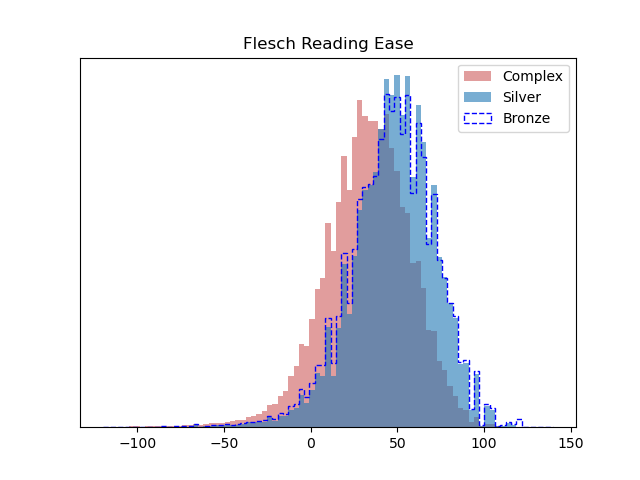}
\end{minipage}
\begin{minipage}[b]{0.32\textwidth}
\centering
\includegraphics[width=\textwidth]{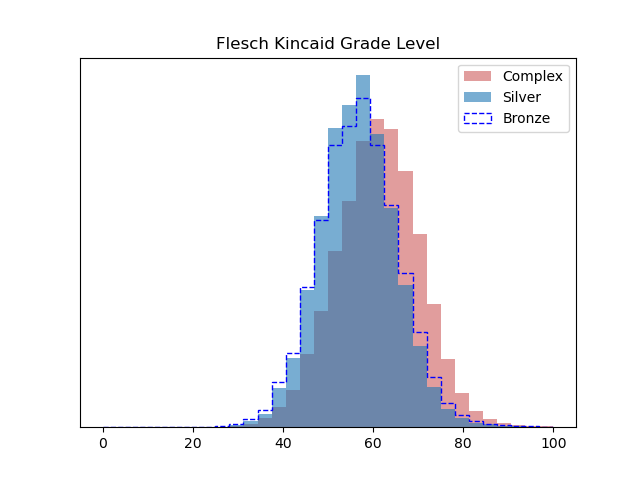}
\end{minipage}
\begin{minipage}[b]{0.32\textwidth}
\centering
\includegraphics[width=\textwidth]{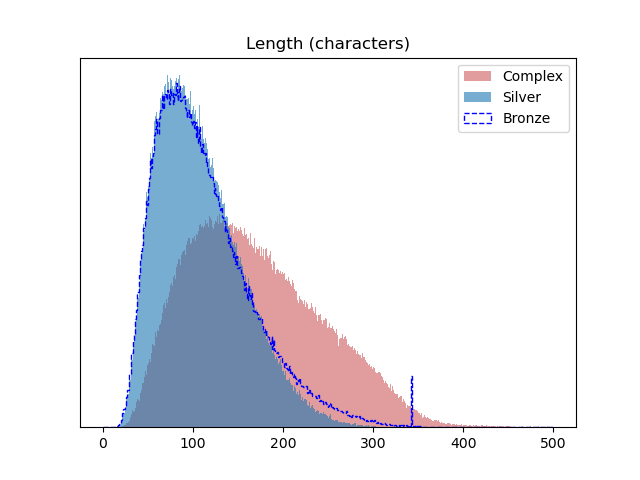}
\end{minipage}
\begin{minipage}[b]{0.32\textwidth}
\centering
\includegraphics[width=\textwidth]{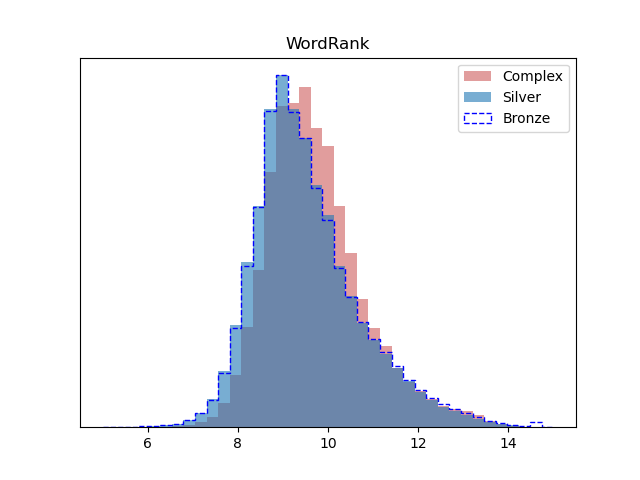}
\end{minipage}
\begin{minipage}[b]{0.32\textwidth}
\centering
\includegraphics[width=\textwidth]{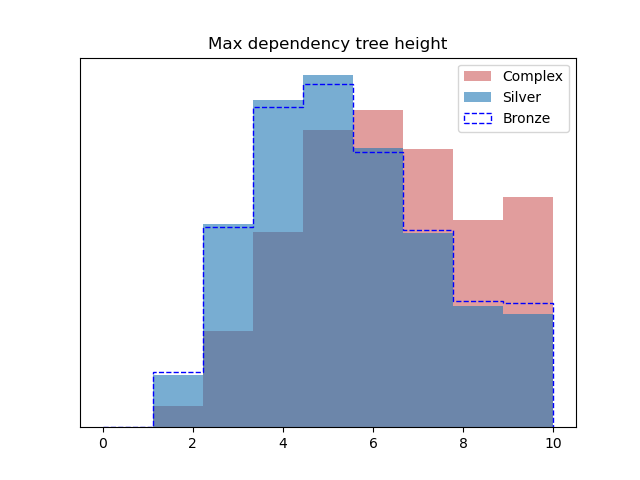}
\end{minipage}
\newline
\centering
\begin{minipage}[b]{0.32\textwidth}
\centering
\includegraphics[width=\textwidth]{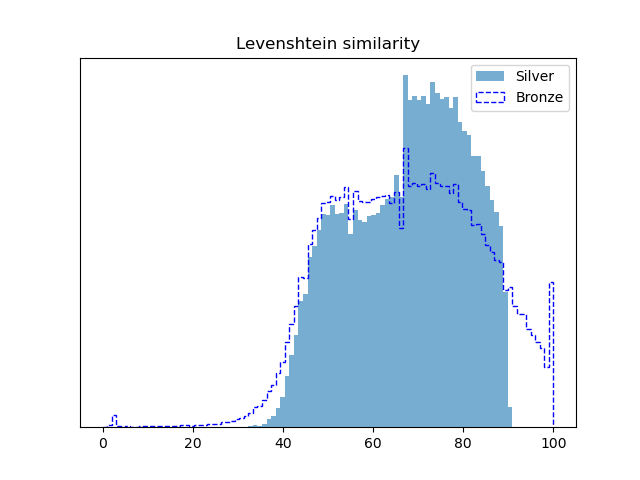}
\end{minipage}
\begin{minipage}[b]{.33\textwidth}
\centering
\includegraphics[width=\textwidth]{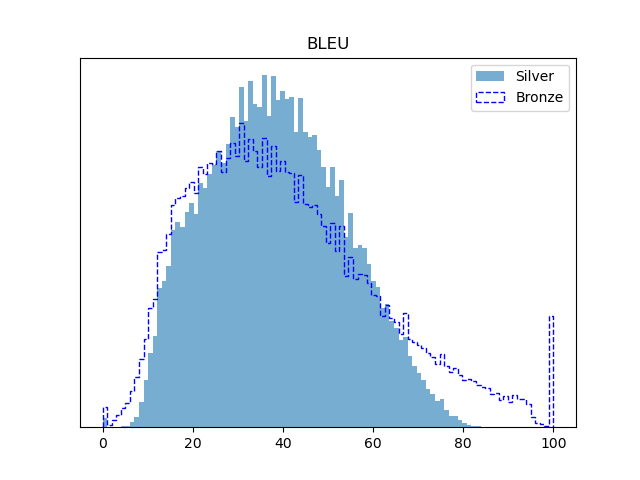}
\end{minipage}
\begin{minipage}[b]{0.32\textwidth}
\centering
\includegraphics[width=\textwidth]{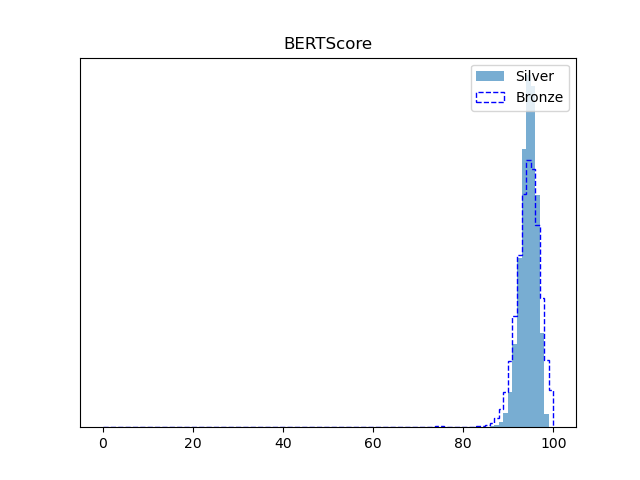}
\end{minipage}
 \caption{Automatic metrics as computed on the complex and simple sentences of the silver corpus. All histograms are normalized to have unit area.}
 \label{fig:results_distribution}

\end{figure*}

To study the dataset characteristics, we compute several automatic metrics. 
In addition to the simplicity proxies discussed in Section \ref{sec:dataset}, we also compute similarity scores among the complex and the simple sentences:
\begin{itemize}
    \item Normalized character-level Levenshtein similarity: it is the number of insertions, deletions, and substitutions between the complex and the simple sentences normalized by the sum of their length. 
    \item BLEU \cite{bleu} and BERTScore \cite{bertscore}: these metrics compute the token-based and semantic similarity of each pair.
\end{itemize}
All metrics are computed per sentence or sentence pairs and then averaged. When required, the original complex sentence is used as the reference.
Figure~\ref{fig:results_distribution} reports the metric distribution; means and standard deviations are summarized in Table~\ref{tab:stats_saentences}.

\begin{table}[]
  \caption{Statistics on sentence pairs and simplified sentences from the silver corpus}
  \label{tab:stats_saentences}
    \centering
    \begin{tabular}{lcc}
    \toprule
    Metric & Complex & Simplified \\
    \midrule
    Flesch Reading Ease \cite{fre} & 33.7 $\pm$ 24.5 &  47.6 $\pm$ 24.2 \\
    Flesch–Kincaid Grade Level \cite{fkgl} & 60.8 $\pm$ 9.1 & 
    56.6 $\pm$ 8.5 \\
    WordRank \cite{access} & 9.7 $\pm$ 1.1 &  9.5 $\pm$ 1.2 \\
    Max dependency tree depth & 6.6 $\pm$ 2.5 & 5.6 $\pm$ 2.2 \\
    Length (chars) & 157.5 $\pm$ 67.4 & 108.4 $\pm$ 46.7 \\
    Levenshtein similarity & \multicolumn{2}{c}{67.7 $\pm$ 12.9} \\
    BLEU \cite{bleu} & \multicolumn{2}{c}{38.6 $\pm$ 15.5} \\
    BERTScore (avg) \cite{bertscore} & \multicolumn{2}{c}{94.45 $\pm$ 1.82} \\
    \bottomrule
    \end{tabular}%
    \end{table}

One can notice that sentences in the silver corpus are simpler than the original ones as measured by the reading scores. The effect on the lexicon is smaller as measured by WordRank. Future work might investigate how to combine our approach with \cite{simplif-backtranslation}, which appears to perform some lexical simplification. The effect of simplification on the syntax complexity is clear as measured by the maximum dependency tree height, and simple sentences are generally significantly shorter than the original. 
Note how the filtering process removes several sentences practically identical to the original one.

\subsection{Qualitative error analysis}
We manually analyzed a subset of 100 sentences to identify the remaining errors. 
Generated sentences are  grammatical and fluent. Errors in meaning preservation are mainly of two types: excessive or wrong compression or problems with compositionality. 
With respect to incorrect sentence compression, coordinate elements in a sentence can be removed. For example, the original sentence 
\textit{"Both the \underline{solid} and the corrugated sheets preferably exhibit on one or both outer sides a layer consisting of the compositions according to the invention.''}
is paired with the sentence 
\textit{"The corrugated sheets have a layer consisting of compositions on one or both sides.''} In this case, the solid sheets, which appear in the original sentence, disappear in the simplified one. 
\noindent Another possible but rare error is the removal of important adjectives. For example, the sentence \textit{"In some cases, it has proved advantageous to use emulsion polymers exhibiting    \underline{reactive} groups at the surface.''} has a simplified pair \textit{"In some cases, it has proved to be beneficial to use emulsion polymers with groups at the surface.''}
Another class of errors derives from the high compositionality of some sentences. In these cases, the relations among elements might not be fully grasped and can be reversed. An example is the following sentence \textit{"The contact lugs, projecting from the bearing plate of the motor, for the electrical feed of the motor are easily accessible there for the connecting cabling in the housing installation orifice.''} which is incorrectly paired with \textit{"There is an easy way to connect the electrical feed of the motor to the housing installation orifice with the contact lugs projecting from the bearing plate.''} 
We also noticed a few hallucinated sentences.
\section{Human evaluation}
\label{sec:human}
\subsection{Evaluation details}
Human evaluation is considered the gold standard when evaluating open-ended tasks \cite{surveynlg, evalSurvey2}. We recruited workers for human evaluation through Prolific\footnote{\url{https://www.prolific.co/}} \cite{prolific} to quantify the quality of the silver standard as judged by non-experts of the patent domain. 
Prolific has been used for human evaluation and annotation, and previous work has found results are more reliable in the analyzed cases than those obtained with Amazon Mechanical Turk \cite{Peer2022}; it is also often regarded as having higher ethical standards. The survey interface was built using Qualtrics\footnote{\url{https://www.qualtrics.com/}}. 

The survey consisted of a statement describing the goal of the study and the data collection process and asking for consent. 
Subjects were required to be at least 18 years old and competent to give consent. The workers were informed that the data were collected for research purposes only but that they would be made available to other researchers and could be used for published work in scientific venues. Workers were required to be native English speakers. 

After a brief demographic section, workers were introduced to the task.
Following previous work, we asked workers to judge the grammaticality and core meaning preservation (adequacy) on a 0-5 Likert scale \cite{likert}. Simplicity was measured in a -2 (the Simplified sentence is much more complex than the original one) to +2 (the Simplified sentence is much simpler than the original one) scale. 
Subjects were also asked to provide an overall 0-5 score and to write a simplification of the original sentence that they considered adequate. 
Numerical values had to be chosen through sliders. 
Instructions included a description of the meaning of Grammaticality, Core Meaning Preservation, and Simplicity, together with clarifications (e.g., regarding judging grammaticality by looking at the Simplified sentence only).
We excluded subjects who did not conclude the survey or who did not report adequate simplifications.
\begin{figure}[]
\centering
\label{fig:survey}
\centering
\includegraphics[width=0.5\textwidth]{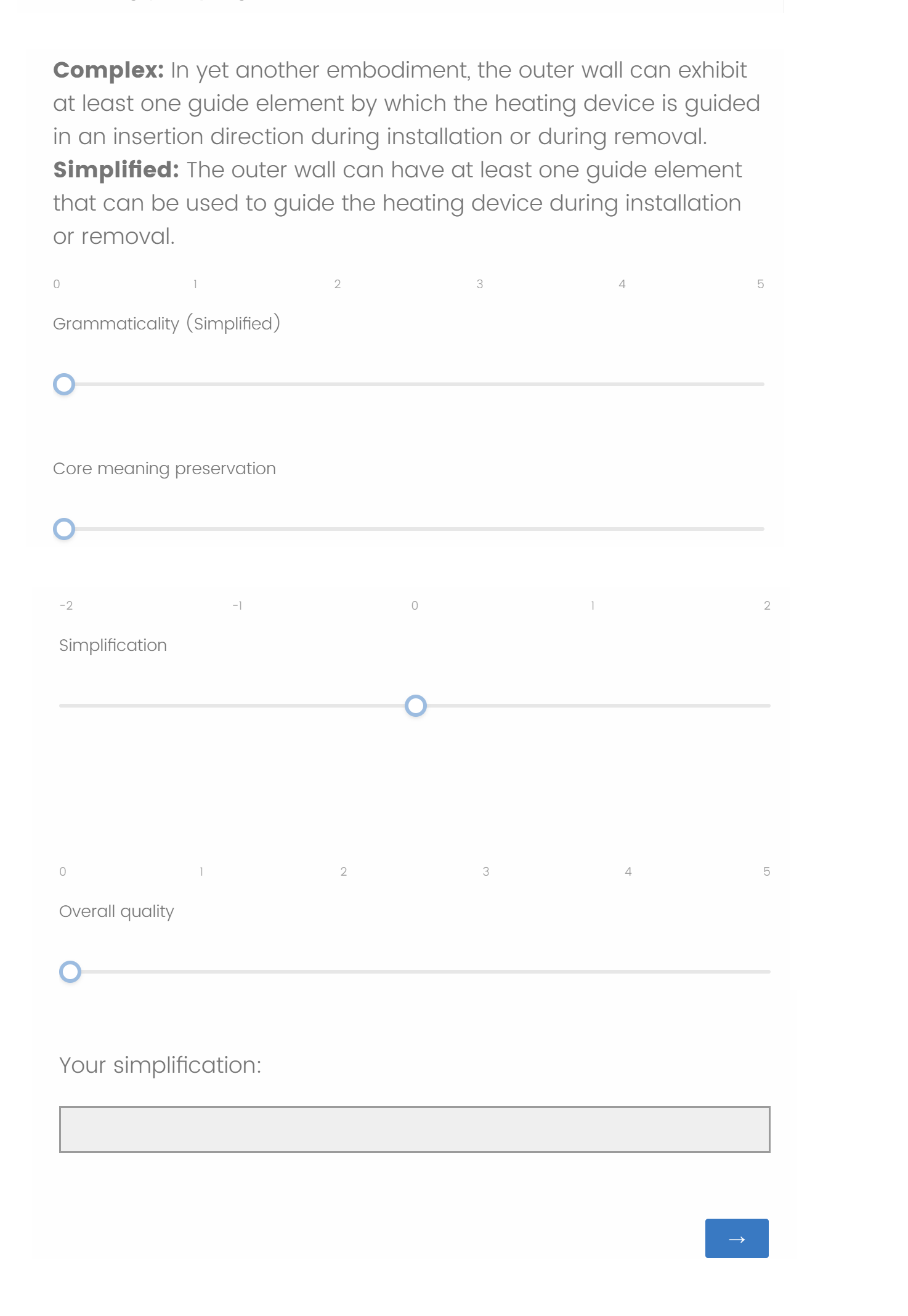}
 \caption{Human evaluation interface. For each sentence, we also included instructions at the top of the page.}
\end{figure}
After a small pilot study and after filtering inadequate responses, we collected evaluations for 96 sentence pairs. 78 workers participated in the study, so each sentence pair was evaluated, on average, by 3.93 participants. 
We chose to evaluate the silver standard with laypeople as they are the target of our simplification. In future work, we plan to further evaluate the dataset with experts of the patent domain and the target technical domain, which might be more reliable, particularly when judging the meaning preservation. 

\subsection{Numerical scores}
\begin{figure}[]
\centering
\begin{minipage}[b]{0.235\textwidth}
\centering
\includegraphics[width=\textwidth]{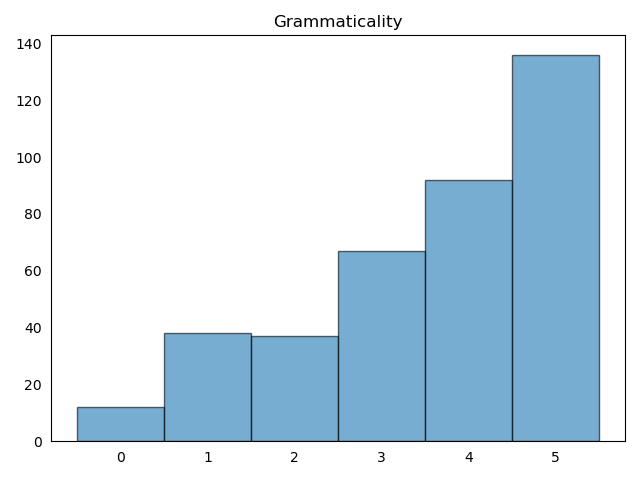}
\end{minipage}
\begin{minipage}[b]{0.235\textwidth}
\centering
\includegraphics[width=\textwidth]{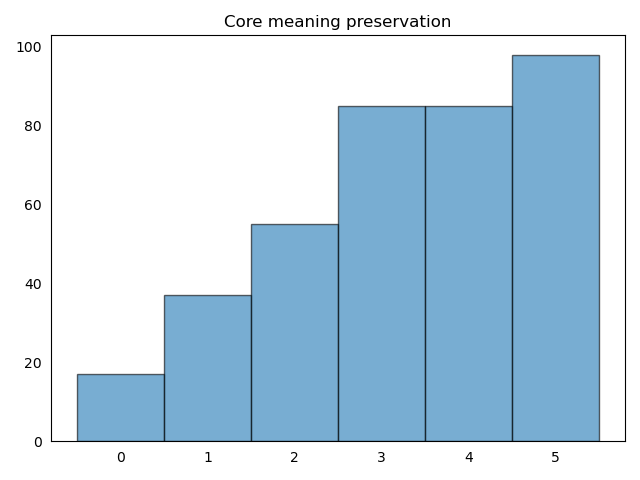}
\end{minipage}
\begin{minipage}[b]{0.235\textwidth}
\centering
\includegraphics[width=\textwidth]{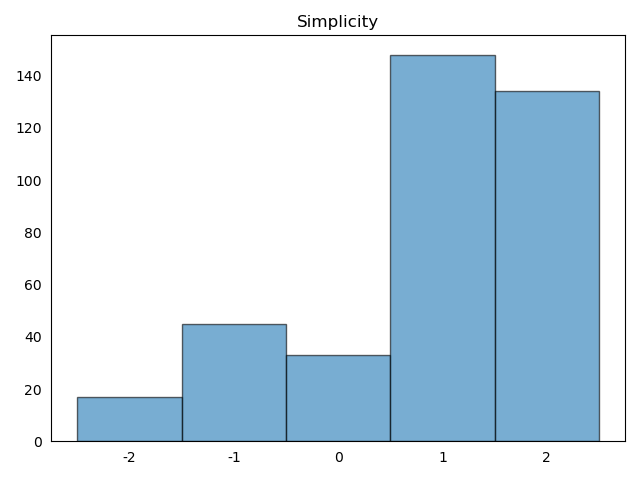}
\end{minipage}
\begin{minipage}[b]{0.235\textwidth}
\centering
\includegraphics[width=\textwidth]{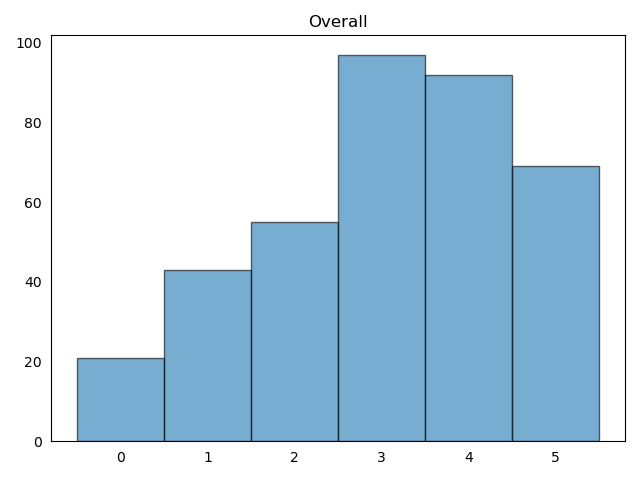}
\end{minipage}
 \caption{Distribution of votes for each dimension. The y-axis shows the absolute number of workers who assigned a given score (x-axis) to sentence pairs in the corpus.}
 \label{fig:human_histo}
\end{figure}

Table \ref{tab:results_ratings} reports the results of the scores as assigned by the Prolific workers. Figure \ref{fig:human_histo} reports the answer distribution for each dimension. 
Results show that sentences are considered rather grammatical. Grammatical sentences were described in the instructions as "free of typos, grammatical or syntactical errors" and workers were asked not to consider style when evaluating this dimension. However, by manually validating the answers, we noticed that complex sentences and sentences containing infrequent grammatical constructs (which are, however, frequent in patent documents) might have a low grammatical score despite no obvious errors.
Most Simplified sentences were considered to retain the original core meaning. 
Regarding simplicity, the vast majority of the sentences were considered somewhat or much simpler than the original.

As a first preliminary experiment, we investigated how each dimension correlated with the overall quality score. 
A simple linear model ($R^2 = 0.66$) scores the dimensions as:
$$ \text{overall} = 0.51 + 0.20 \times \text{G} + 0.45 \times \text{M} + 0.40 \times \text{S} $$
where G is the grammaticality score, M is the core meaning score and S is the simplicity score.

\begin{table}[]
  \caption{Human evaluation results}
  \label{tab:results_ratings}
    \centering
    \begin{tabular}{ll}
    \toprule
    Dimension & Mean $\pm$ std \\
    \midrule
    Grammaticality [0,5] & 3.59 $\pm$ 1.45
    \\
    Core meaning preservation [0,5] & 3.27 $\pm$ 1.46 \\
    Simplicity [-2,2] & 0.89 $\pm$ 1.15 \\
    Overall [0,5] & 3.07 $\pm$ 1.44 \\
    \bottomrule
     \end{tabular}%
\end{table}

\subsection{Human-written simplifications}
\label{sec:human_annotation}

\begin{table*}[]
\label{tab:simpl_humans}
\caption{Examples of human simplifications}
\begin{tabular}{lll}
    \toprule
    \footnotesize{Original} & \footnotesize{Simplified (automatic)} & \footnotesize{Simplified (human)} \\
    \midrule
     \parbox{5cm}{\footnotesize{In yet another embodiment, the outer wall can exhibit at least one guide element by which the heating device is guided in an insertion direction during installation or during removal.}} & \parbox{5cm}{\footnotesize{The outer wall can have at least one guide element that can be used to guide the heating device during installation or removal.}} & 
     \parbox{5cm}{\footnotesize{The outer wall will have at least 1 guide to help guide the heating device during installation or removal. \newline
     During installation or removal the outer wall has 1 or more guide element.}} \\
     \\
     \parbox{5cm}{\footnotesize{A lamp holder or fitting carrier for at least one lamp is arranged in a position in which it lies outwardly opposite a recess in the reflector and is releasably connectable with the base frame by second mounting means.}} & \parbox{5cm}{\footnotesize{A lamp holder or fitting carrier for at least one lamp is arranged in a way that it is releasably connectable with the base frame by second mounting means.}} & 
     \parbox{5cm}{\footnotesize{A lamp holder or fitting carrier for at least one lamp is arranged in a way that it is connectable with the base frame by second mounting means. \newline
     A lamp holder or fitting carrier for at least one lamp is arranged in a position releasably connectable with the base frame by second mounting means. \newline
     A lamp holder is arranged in a way that lets the base frame be accessible for a second mount. \newline
     this product is easily changed from a one lamp to a 2 lamp item via the changeable base}}\\
     \\
    \parbox{5cm} {\footnotesize{In an emergency, the safety valve can also be opened by hand, when the pressure in the feed line is too high for unknown reasons.}} & \parbox{5cm}{\footnotesize{When the feed line is too high for unknown reasons, the safety valve can be opened by hand.}} & 
    \parbox{5cm}{\footnotesize{In an emergency, the safety valve can be opened by hand, when the pressure is too high \newline When the feed line pressure is too high, the safety valve can be opened by hand. \newline In an emergency, the feed line is too high for unknown reasons, the safety valve can be opened by hand. \newline when feed line is too high open safety valve \newline if the line feed is too high, hand open the valve}} \\
    \bottomrule
    \end{tabular}%
\end{table*}

During the survey, we also asked participants to provide a simplification, in their own words, of the original sentence. Table \ref{tab:human_sentences} shows some examples generated by the workers.
Table 
\ref{tab:human_sentences} reports some metrics computed on the collected data. 
Unsurprisingly, the human-written sentences are more similar to the simplified sentence; note, however, that their similarity is not as high as to consider them derivative. Moreover, they are shorter than synthetic simplified sentences.
Considering the simplification scores, they are in line with that generated by the model. The only exception is that of the dependency tree, which is sensibly less deep in the case of sentences generated by humans.
Human-written simplifications allow evaluating models trained on our silver standard with data that are not synthetic. However, users should be aware of their limitations, as they were produced by workers lacking expertise in the patent or technical domain.

\begin{table}[]
  \caption{Statistics on simplified sentences produced by humans }
  \label{tab:human_sentences}
    \centering
    \begin{tabular}{lcc}
    \toprule
    Metric & Complex & Simplified \\
    \midrule
    Levenshtein similarity  & 58.87 $\pm$ 19.47 & 65.98 $\pm$ 20.13 \\
    BLEU \cite{bleu} & 23.48 $\pm$ 19.39 & 35.56 $\pm$ 26.47 \\
    BERTScore (avg) \cite{bertscore} & 92.15 $\pm$ 4.03 & 93.58 $\pm$ 4.11 \\
    Flesch Reading Ease \cite{fre} & \multicolumn{2}{c}{45.92 $\pm$ 29.91} \\
    Flesch–Kincaid Grade Level \cite{fkgl} & \multicolumn{2}{c}{56.40 $\pm$ 11.83} \\
    WordRank \cite{access} & \multicolumn{2}{c}{9.47 $\pm$ 1.46} \\
    Max dependency tree depth & \multicolumn{2}{c}{4.85 $\pm$ 2.02} \\ 
    Length (chars) & \multicolumn{2}{c}{88.67 $\pm$ 41.96} \\
    \bottomrule
    \end{tabular}%
    \end{table}

\section{Conclusions}
\label{sec:conclusion}
In this paper, we have discussed a method to generate a parallel silver standard for simplifying patent sentences through rephrasing. 
To the best of our knowledge, this is the first parallel simplification corpus for patents. 
We have analyzed the corpus quantitatively and qualitatively, showing that it can be used to train a sequence-to-sequence simplification model. We have also conducted a large-scale human evaluation of the corpus and collected human-written simplifications. 
While not exempt from the shortcoming of its automatic origin, we have shown that filtering out faulty candidate allows us to obtain a corpus that has been considered grammatical, adequate, and with a significant simplification.
In future work, we plan to explore alternative models to generate the simplification candidates. 
We will also further explore whether transfer-learning techniques can further improve results. 
Finally, we plan to investigate whether the method exposed in this paper can be successfully adopted for other types of technical text.

\section*{Acknowledgments}
This work is (partly) supported by the Spanish State Research Agency under the Maria de Maeztu Units of Excellence Programme (CEX2021-001195-M) and the
Context-aware Multilingual Text Simplification (ConMuTeS) project  PID2019-109066GB-I00/AEI/10.13039/ 501100011033 awarded by Ministerio de Ciencia, Innovación y Universidades (MCIU) and by Agencia Estatal de Investigación (AEI) of Spain. The authors also thank the anonymous reviewers and Kim Cheng Sheang for his help and suggestions.

\bibliographystyle{unsrt}  
\bibliography{references}

\end{document}